\documentclass{article}

\usepackage{todonotes}

\usepackage[utf8]{inputenc}
\usepackage[english]{babel}
\usepackage[T1]{fontenc}
\usepackage{amsmath}
\usepackage{amsfonts}
\usepackage{amsthm}
\usepackage{amssymb}
\usepackage{placeins}
\usepackage{enumitem}
\usepackage{algpseudocode}
\usepackage{algorithm}
\usepackage{graphicx}
\usepackage{url}
\usepackage{listings}
\usepackage{setspace}
\usepackage{tabularx}
\usepackage{tablefootnote}
\lstdefinestyle{mystyle}{
    numberstyle=\small,
    numbers=left,
    breaklines=true,                 
    captionpos=b,
    keepspaces=true,
    numbersep=10pt,                  
    columns=flexible,
    escapeinside={(*@}{@*)}
}
\usepackage{xspace}
\usepackage{xcolor}
\usepackage{enumitem}
\usepackage{longtable}
\usepackage[caption=false,font=normalsize,labelfont=sf,textfont=sf]{subfig}
\usepackage{wrapfig
}
\usepackage[acronym]{glossaries}
\usepackage{authblk}

\newfloat{lstfloat}{htbp}{lop}
\floatname{lstfloat}{Listing}

\newcommand{\gavelowl}{\textit{Gavel-OWL}}
\newcommand{\gavel}{\textit{Gavel}}

\title{When one Logic is Not Enough: Integrating First-order Annotations in OWL Ontologies}

\begin{document}
\author[1]{Simon Flügel}
\author[1]{Martin Glauer}
\author[1]{Fabian Neuhaus}
\author[2,3]{Janna Hastings}
\affil[1]{Institute for Cooperating Systems, {Otto von Guericke University Magdeburg},
{Germany}
}
\affil[2]{Faculty of Medicine, Institute for Implementation Science in Health Care, {University of Zurich}, {Switzerland}
}
\affil[3]{School of Medicine, {University of St Gallen}, {Switzerland}
}

\maketitle


\begin{abstract}
In ontology development, there is a gap between domain ontologies which mostly use the web ontology language, OWL, and foundational ontologies written in first-order logic, FOL.
To bridge this gap, we present Gavel, a tool that supports the development of heterogeneous 'FOWL' ontologies that extend OWL with FOL annotations, and is able to reason over the combined set of axioms.
Since FOL annotations are stored in OWL annotations, FOWL ontologies remain compatible with the existing OWL infrastructure.
We show that for the OWL domain ontology OBI, the stronger integration with its FOL top-level ontology BFO via our approach enables us to detect several inconsistencies.
Furthermore, existing OWL ontologies can benefit from FOL annotations. 
We illustrate this with FOWL ontologies containing mereotopological axioms that enable new meaningful inferences. 
Finally, we show that even for large domain ontologies such as ChEBI, automatic reasoning with FOL annotations can be used to detect previously unnoticed errors in the classification.
\end{abstract}


%
%
%
%

\section{Introduction}
\label{sec:introduction}
The landscape of applied ontology contains a rift. The vast majority of ontologies that are used in computational systems are written in OWL, more specifically OWL 2 DL or OWL language profiles, which are essentially sublanguages of OWL 2 DL.\footnote{In the following we will for the sake of brevity refer to OWL 2 DL and OWL 2 language profiles as `OWL'. We are aware of  so-called `OWL 2 Full' (i.e., OWL 2  without DL constraints  and a RDF-based semantics), but since it is rarely used, we will not consider it in this paper. }
However, the authors who work on the foundational topics of applied ontology, such as upper level entities and their axiomatisations, typically do not use OWL. E.g., in 2020 and 2021 the \emph{Formal Ontology in Information Systems} conferences accepted altogether 19 papers on foundational topics. Of those, none used OWL as representational language. 

The reason for this language rift is easy to explain. The OWL Manchester Syntax is relatively easy to learn for domain experts, and there are an increasing number of widely available tools supporting development of ontologies in the OWL language. Furthermore, OWL is based on the description logic SROIQ and, thus, satisfiability of OWL ontologies and logical inference with OWL ontologies are decidable decision problems. 
Both factors are major advantages for anybody who develops a domain ontology in cooperation with domain experts and intends to use automatic reasoning (e.g., using consistency checks for quality control). 
However, the advantages come with a price: in order to improve readability and achieve decidability one needs to sacrifice expressivity. 
Consequently, many ontological differences that are studied by foundational ontologists are not expressible in OWL.  
E.g., part-whole relationships are a staple of most ontologies and have been extensively studied by foundational ontologists. However, in OWL 2 DL it is even impossible to axiomatise proper parthood as strict order, let alone represent interesting distinctions  between different mereologies \cite{keet2012ontoparts}.
Hence, foundational ontologists typically use first-order logic (FOL) or even more expressive logics as representational languages. 
Consequently, the vast majority of papers on foundational ontology contain axioms and definitions that cannot be directly reused by the ontologies that are used in practice in information systems. 
The language rift thus prevents theoretical results and insights from achieving their full impact in benefiting downstream practical applications. 

A particularly interesting case are upper level ontologies, which aim to provide a reusable framework for the development of domain ontologies. There are a number of different upper level ontologies \cite{borgo2022Foundational}, which reflect a wide range of philosophical points of view. Most widely used is the Basic Formal Ontology (BFO), which is used by more than 250 ontology projects. BFO is grounded in a rich Aristotelian tradition, is described in numerous publications, a user guide, and a textbook, and most recently it has even been standardised by the International Organization for Standardization as ISO/IEC 21838-2:2021. However, the version of BFO that is actually used in most information systems is BFO 2.0 OWL, which contains exactly 52 axioms, more precisely  18 disjointness axioms and 34 subsumption axioms between atomic classes. 
Interestingly, BFO 2.0 OWL contains annotations that include additional axioms in the Common Logic Interchange Format (CLIF), a variant of FOL\footnote{Technically, Common Logic is more expressive than FOL \cite{iso2018commonlogic}, but for the purpose of this paper the differences may be ignored.}, which provide a much richer logical axiomatisation than the OWL axioms.  However, since OWL reasoners ignore annotations, these axioms are computationally inert; they only serve as documentation for the human readers.  
In summary, while there is a rich literature on BFO, 
almost none of it is  materialised in the format that is actually used by domain ontologies.  From a logical point of view BFO 2.0 OWL is just a simple taxonomy with a few  additional disjointness axioms. 
Consequently, the developers of domain ontology  cannot use OWL reasoners or other automatic reasoners to check whether their ontology actually conforms to BFO.

The goal of this paper is to present {\gavel}~\footnote{https://github.com/gavel-tool/python-gavel} and its OWL-specific extension {\gavelowl}~\footnote{https://github.com/gavel-tool/python-gavel-owl}, tools that were developed to bridge the language rift by enabling ontology developers to  develop heterogeneous ontologies that contain both FOL and OWL axioms, 
and supports reasoning with the integrated model.  
One major benefit of {\gavel} and {\gavelowl} is that it does not require 
developers of applied ontologies to make any changes to their established workflow for OWL ontologies: developers can continue to use the same tools (e.g., Protégé) to develop their ontologies, and use standard OWL reasoners to reason with the OWL part of the ontology. {\gavelowl} merely offers the additional option to enhance an OWL ontology with FOL axioms. These additional axioms may be based on a foundational ontology, but they may also be domain specific axioms that are expressible in FOL but not in OWL. 

In section~\ref{sec:requirements} that follows, we discuss the requirements for {\gavelowl} and our approach for addressing them.  The capabilities of {\gavelowl} and its implementation are presented in section~\ref{sec:capabilities}. Afterwards, we discuss in section~\ref{sec:evaluation} the merits of the approach by first showing how {\gavelowl} overcomes some limitations of OWL that have been discussed in the literature, and then we consider two case studies. 
In the first, we will show that BFO CLIF axioms may be used to find errors in an BFO-based OWL domain ontology. 
In the second, we extend a large domain ontology with FOL axioms and illustrate how automatic theorem proving with the heterogeneous ontology yields new subsumptions that an OWL reasoner on its own is not able to detect. In the remainder of the paper we discuss related work, the significance and potential impact of the approach. 




\section{Requirements and Approach}\label{sec:requirements}
Our goal is to enable ontology developers to benefit from the advantages of OWL, including its user friendly syntax, decidability, and mature tool chain, without having to forego the ability to use the greater expressivity of FOL when it is required, or when a preexisting FOL ontology is available. 

Thus, a \textit{FOWL ontology} consists of two components, an OWL component and a FOL component, which extends the OWL axioms with additional axioms in FOL.  
To achieve this goal our solution ideally should meet the following requirements: 
\begin{enumerate}
	\item   In order to remain accessible by the existing OWL infrastructure, FOWL ontologies should be syntactically valid OWL ontologies. 
	\item   The OWL component of a FOWL ontology should be usable by existing OWL tools and follow the OWL 2 DL semantics as specified in \cite{owl2directsemantics}. 
	\item   For convenience of editing and debugging, it should be possible to associate FOL axioms closely with related OWL entities and axioms. Ideally, this should be able to be done with tools like Protégé, which are widely used for ontology editing. 
	\item In addition, it should be possible to integrate existing FOL ontologies (e.g., a foundational ontology) with a FOWL ontology.  
 
	\item Automatic reasoning with an FOWL ontology should utilise both its OWL and its FOL axioms. 
\end{enumerate}

Our approach for meeting these requirements is to create a combined OWL and FOL ontology  through annotations in OWL ontologies. Thus, in some sense we follow the precedent set by BFO 2.0 OWL, which is also an OWL file that contains FOL annotations. However, while in BFO 2.0 OWL the FOL annotations are logically inert, {\gavelowl} enables automatic reasoning with the logical theory that is the result of combining the OWL axioms with the FOL axioms. 

%
More specifically, a \textit{FOWL ontology} is an OWL file, where 
one or multiple annotation properties are selected, and all values of \textit{AnnotationAssertion} axioms with these annotation properties are interpreted as FOL axioms.
Hence, the heterogenous ontology is represented in a syntactically valid OWL file and the OWL part of the ontology may be processed by standard OWL tools, because the FOL annotations do not influence the ontology's OWL semantics.

\begin{figure}
    \includegraphics[width=0.6\textwidth]{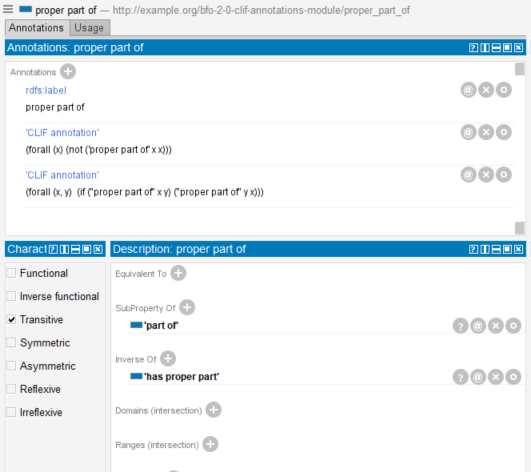}
	\centering
	\caption{View of an annotated OWL entity in Protégé.}
	\label{fig:methodlogy_protege}
\end{figure}

As they are stored in annotation values, the FOL axioms also remain closely connected to the OWL axioms: typically, all axioms for a given subject entity are grouped together.
For instance, Figure~\ref{fig:methodlogy_protege} shows how Protégé  displays annotations and logical OWL axioms for the entity \textit{proper part of}.
The FOL axioms can be seen in the upper part and the OWL axioms in the lower part.
Thus, a user is able to view related FOL and OWL axioms on the same screen. 

{\gavelowl} supports two different syntaxes for FOL annotations: CLIF and TPTP. CLIF is a human readable syntax for ISO Common Logic \cite{iso2018commonlogic}.\footnote{Strictly speaking, CLIF's syntax is more flexible than `normal' FOL because it does not distinguish between individual constants, predicate and function symbols. In addition, CLIF includes sequence variables which increase the expressiveness of CLIF beyond FOL.  However, for the purpose of this paper we treat CLIF just as an alternative syntax for `normal' FOL. }    
The TPTP syntax is widely used in the automatic theorem proving community, in particular the TPTP problem library \cite{sutcliffe2017tptp, sutcliffe2006tptp}.
To increase the readability of TPTP, we do not use complete TPTP axioms, but only their formula part. E.g., 
 we write 
 \begin{verbatim}
     ![X] ~ 'proper part of' (X,X)
 \end{verbatim}
instead of the full TPTP expression 
\begin{verbatim}
    fof(axiom_name, axiom, (![X] ~ 'proper part of' (X,X))).
\end{verbatim}

Reasoning with a FOWL ontology requires the integration of its OWL component with its FOL component. This is done by translating the OWL axioms into FOL (in TPTP syntax) and integrating them with the FOL axioms from the annotations\footnote{CLIF axioms are also translated into TPTP syntax.} as well as, potentially, additional axioms from a background theory. The result is a FOL ontology in TPTP syntax, which can be used for automatic reasoning and consistency checking. To enable these automated consistency checks, {\gavel} is able to interact with the theorem prover Vampire \cite{kovacs2013vampire}. Furthermore, {\gavel} features a dynamic extension scheme that allows for the integration of other logics through new plugins. We present here {\gavelowl}, which allows the integration of first-order logic and OWL. Other ontologies may also use logics for annotation, such as typed logics or higher order logics. These can then be connected to {\gavel} by separate plugins similar to {\gavelowl} and to use the same proof pipelines described in this submission.

One obstacle for the integration of OWL axioms and FOL annotations is that it is necessary to map their signatures. 
E.g., in the example in Figure~\ref{fig:methodlogy_protege} an OWL entity, more specifically an 
 object property,  is annotated with a CLIF axiom.  The OWL and the FOL axioms together specify proper parthood as strict partial order. (This could not be axiomatised in OWL 2 DL, since the axiom would violate its global restrictions.) 
However, strictly speaking, the OWL axioms are about the entity uniquely identified by a long IRI 
 (namely, \url{http://purl.obolibrary.org/obo/BFO_0000175}) ,  which is then annotated with the literal 'proper part of'\textasciicircum\textasciicircum{xsd:string} as label, while the CLIF axioms are using neither the OWL IRI nor the exact label of the OWL entity, but the CLIF name ``proper\_part\_of". This is indicative of a typical problem. IRIs are often too long to be conveniently typed by human. Thus, one cannot expect the FOL annotations to include the complete IRIs that are used to identify an OWL entity.  
 Further, following recommended best practices \cite{mcmurry_identifiers_2017}, many OWL ontologies contain IRIs that consists of arbitrary alpha-numeric identifiers, thus, they are not human readable. In addition, CLIF and TPTP have both restrictions on their syntax. Hence, typically, there are no exact matches between the signatures of the OWL ontology and the signature of the FOL axioms. {\gavelowl} attempts to automatically resolve these issues by considering suffixes of IRIs and the labels of OWL entities, when trying to match OWL entities to FOL symbols.

All in all, our methodology provides a practical, user-friendly approach to combining the advantages of both OWL and FOL:
By keeping the main part of the ontology in OWL and only adding FOL axioms as annotations, the ontology can be used as a standard OWL ontology.
This means, that developers can make use of the existing tools, reasoners and their familiarity with the language.
The FOL annotations can then be added to extend the ontology with axioms that are not expressible in OWL, increasing the possibilities developers have for modelling their domain as accurately as possible and allowing the integration of FOL upper level ontologies.
Reasoning support for the complete ontology is ensured by {\gavelowl} via translating into a unified FOL theory.

\section{Capabilities and Implementation}
\label{sec:capabilities}

To achieve its objective of supporting the development of heterogeneous ontologies that consist of an OWL file with FOL annotations (either in CLIF or TPTP syntax), {\gavelowl} has the following capabilities:
\begin{enumerate}
	\item OWL ontologies can be translated into a file in TPTP syntax.  
	\item OWL ontologies that are annotated with FOL axioms (in either CLIF or TPTP) syntax can be translated into a file in TPTP syntax, which integrates the translation of the OWL axioms with the FOL axioms in the annotations. The signatures are harmonised, optionally using IRIs or human friendly identifiers.
	\item OWL ontologies that are annotated with FOL axioms can be translated into the Distributed Ontology, Modelling and Specification language (DOL), where the OWL component and the FOL component of the heterogeneous ontology are represented as two different sub-ontologies. 
	\item OWL ontologies  can be checked for consistency via the OWL reasoner  HermiT~\cite{glimm2014hermit}. 
	\item An automatic theorem prover for FOL (currently, Vampire)  may be used to attempt to prove FOL conjectures (in TPTP syntax) from an OWL ontology  (possibly with FOL annotations).\label{fol-conjectures} 
	\item Alternatively to (\ref{fol-conjectures}), the conjectures may also be provided by a second OWL ontology. In this case {\gavelowl} attempts to prove that the second ontology is logically entailed by the first. This is implemented by translating the axioms of the second ontology into FOL and using them as conjectures of TPTP problems.  
\end{enumerate}

The {\gavelowl} implementation builds on two pre-existing components.
First, the \textbf{OWL API}~\cite{horridge2011owlapi}~\footnote{https://github.com/owlcs/owlapi} is an open source Java library for working with OWL ontologies.  
It has originally been developed by the University of Manchester and is currently maintained by Matthew Horridge and Ignazio Palmisano. We use it for parsing OWL ontologies and extracting FOL annotations. Since the OWL API is a Java library, {\gavelowl} includes a \textit{Java Server} that handles the interaction with the OWL API and contains large parts of the OWL-related logic. It is connected to the application's Python part via Py4J~\footnote{https://www.py4j.org/}.
%
Second, \textbf{Macleod}~\footnote{https://github.com/thahmann/macleod} is a Python tool for working with ontologies written in the CLIF syntax, created by Torsten Hahmann.
We use it to translate CLIF annotations into TPTP syntax. 
%
%

\begin{figure}
    \includegraphics[width=.9\textwidth]{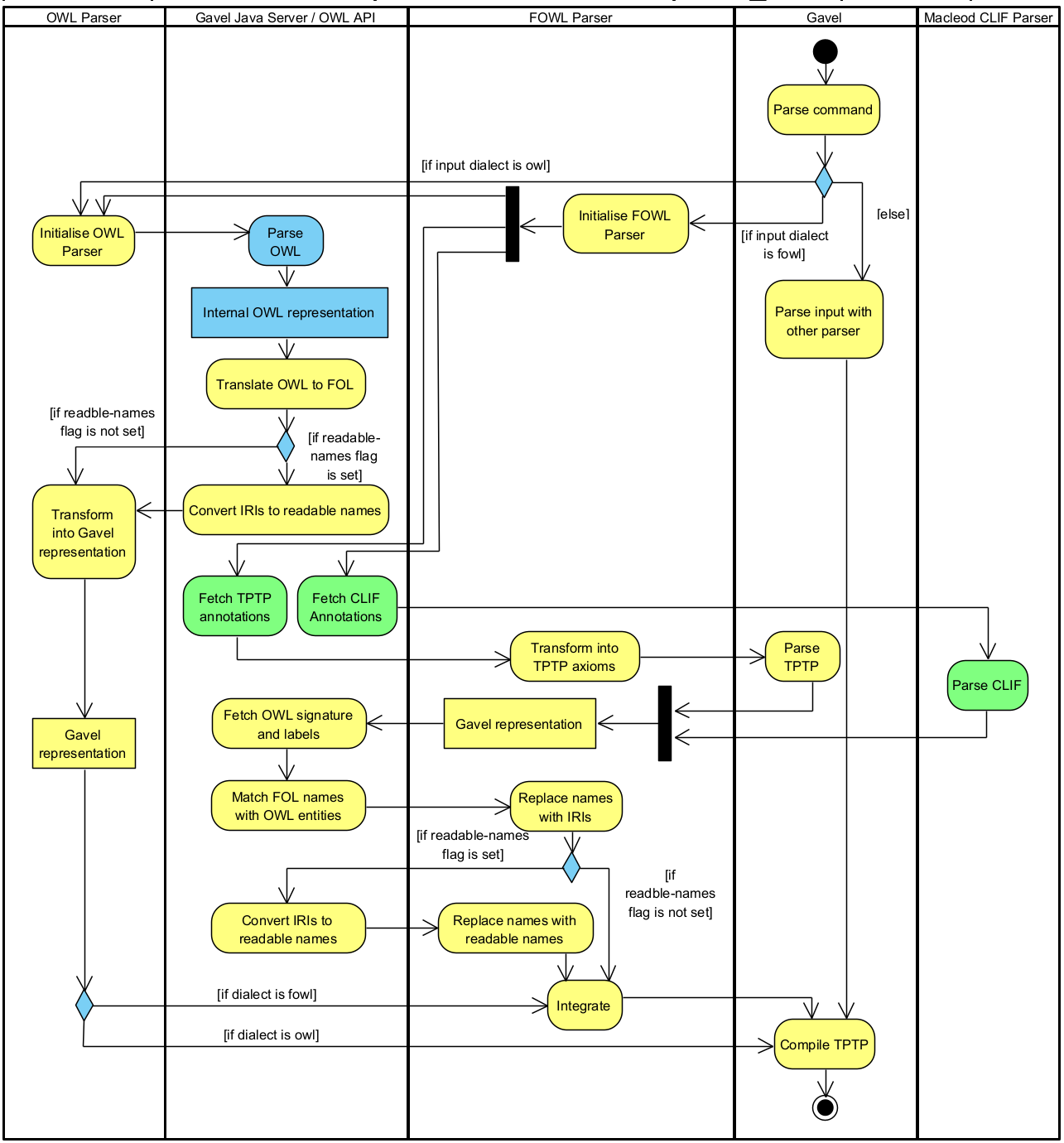}
	\centering
	\caption{A UML activity diagram illustrating the process of translating an annotated OWL ontology or standard OWL ontology into TPTP. Actions that were newly implemented for the translation tasks are shown in yellow, actions that were reused from existing tools are marked in blue. For green actions, the existing functionality has been extended.}
	\label{fig:fowlactivitydiagram}
\end{figure}

Figure~\ref{fig:fowlactivitydiagram} illustrates the implementation of the main functionality of {\gavelowl}, namely the translation of an OWL ontology (possibly with FOL annotations) to FOL.
In the case of an OWL ontology without any FOL annotations the following things happen: 
First, it parses the OWL ontology by using the OWL API and subsequently translates it into FOL. 
Optionally, IRIs are replaced with more readable names.
The translation from OWL to FOL, previously described in \cite{flugel2021fowl}, is based on the OWL Direct Semantics~\cite{owl2directsemantics} and uses a mapping $[ \quad ]^{p}_q$ between OWL and FOL expressions.\footnote{A detailed specification of the mapping is available at \url{https://github.com/gavel-tool/python-gavel-owl/wiki/OWL-to-FOL-mapping}.}
This mapping is defined for each basic type of OWL axiom or expression and is then applied recursively on more complex axioms. In doing so, the parameters $p$ and $q$ are used to keep track of substitutions that have to be made and are left out if not needed.
E.g., $\mathit{SubclassOf}(\mathit{Fish}, \mathit{Animal})$ would be translated as follows. 

\vspace{-2em}
$$[\mathit{SubclassOf}(\mathit{Fish}, \mathit{Animal})]
\quad \Leftrightarrow \quad \forall x ([\mathit{Fish}]^x \rightarrow [\mathit{Animal}]^x)
\quad \Leftrightarrow \quad \forall x (\mathit{Fish}(x) \rightarrow \mathit{Animal}(x))$$
\vspace{-2em}

%
%
A FOL translation of a complete OWL ontology consists of the translations for each axiom combined with additional background axioms. These  capture general assumptions of the Direct Semantics (e.g., the distinction between object and data domains) and the meaning of the OWL reserved vocabulary, such as \texttt{owl\!:\!Nothing}.
These steps are implemented in the Java Server, because they build upon the OWL API functionality.
Therefore, the resulting FOL axioms need to be transferred back to the Python OWL Parser and transformed into Gavel's internal representation.
This representation is then used by Gavel to compile it into a TPTP document, which gets returned to the user.

OWL files with FOL annotations are split into three parts. 
The OWL part of the ontology is translated into the internal FOL representation of Gavel as described above. 
TPTP annotations are extracted,  turned into valid TPTP expressions and then parsed by Gavel.  
CLIF is parsed with the help of Macleod, which has been modified to return Gavel's internal representation as well. 
In order to align the FOL signature with the OWL signature, the Java Server first fetches the OWL signature and \textit{rdfs:label}-values and then maps them to the FOL names using the Levenshtein distance. If no close match is available, the name will be used as-is, making the assumption that there is no OWL entity to be found and that this name only appears in FOL axioms.
Based on this mapping FOL names are replaced with IRIs and in a second step, if required, with readable names. Finally, the FOL axioms from annotations are combined with the FOL translation of the OWL part, which also uses either IRIs or readable names, and compiled to TPTP, completing the translation process.

\section{Evaluation and Use Cases}
\label{sec:evaluation}
We aim to evaluate {\gavelowl} in use for ontology development. For this purpose, we first evaluate whether {\gavelowl} helps to overcome OWL restrictions \emph{that are ontologically meaningful}. 
Since {\gavelowl} enables reasoning with arbitrary FOL axioms, it also enables inferences with FOWL ontologies that an OWL reasoner would not be able to support. 
However, it is not obvious whether this additional expressivity is also useful. Thus, we evaluate the capability of {\gavelowl} based on examples from the literature in which the authors propose ontologically meaningful modelling patterns that cannot be expressed in OWL.  

As mentioned in the introduction, one motivation for developing {\gavelowl} is to support a better integration of upper level ontologies with domain ontologies. Thus, as a second way to show the benefits of {\gavelowl}, we developed a new version of BFO that integrates OWL and FOL, and used it to check consistency of the Ontology for Biomedical Investigations (OBI)~\cite{bandrowski2016obi}, a widely used ontology in the life sciences.  

In another case study we show how domain specific  FOL axioms  may be used to 
increase the quality of a domain ontology. For this purpose we extended the  Chemical Entities of Biological Interest (ChEBI) ontology \cite{hastings2016chebi} with roughly 120 000 FOL axioms. 
Gavel allowed us to detect errors in ChEBI, which OWL reasoning alone was not able to detect.   This example illustrates that even large domain ontologies may benefit from FOL axioms.  

In a previous publication \cite{flugel2021fowl} we discussed the performance of the translation from OWL to TPTP (e.g., time 
 per translated axiom). We found that even large ontologies may be translated in a reasonable amount of time (e.g., the Gene Ontology is translated in less than 30 minutes on a normal laptop). However, since there is no reason why one would translate the same large ontology regularly, performance may not be the most important metric. 

\subsection{Evaluation via Ontologically Meaningful Inferences}
For the purpose of this evaluation we consider
modelling patterns that 
have been identified in the literature as ontologically meaningful, but which cannot be expressed in OWL. The first examples are mentioned in a 
paper by Schneider, Rudolph and Sutcliffe on a version of OWL 2 DL without global restrictions~\cite{schneider2012owlnorestrictions}.
In order to show the advantages of their unrestricted OWL version, they provide 12 examples of modelling patterns that are useful but are not syntactically valid OWL 2 DL, since they 
violate its global restrictions. For each example they also provide a conjecture ontology, which  is logically entailed according to the OWL Direct Semantics, but cannot be inferred by OWL reasoners because of the global constraint violation.   
For our purposes we replaced the OWL axioms that violate the DL restrictions with logically equivalent FOL annotations. Thus, the result are twelve pairs of ontologies, each pair consisting of a FOWL premise ontology  representing a useful modelling pattern and a OWL conjecture ontology that is  entailed by the premise ontology, but cannot be inferred by OWL 2 DL reasoners. 
In each of the 12 examples,  {\gavelowl} is   able to prove that  the  conjecture ontology is logically entailed by the FOWL premise ontology. 
%
%

A second set of test cases has been derived from a paper by Keet et al. 
on mereotopological relations~\cite{keet2012ontoparts}, in which the KGEMT mereotopological theory~\cite{aiello2007spatiallogics} is used to strengthen the axiomatisation of ontologies and thereby improve their accuracy.
One of the challenges Keet et al. face is the limited expressiveness of OWL:
Of the 27 definitions and axioms they identified in KGEMT only 9 can be fully represented in OWL 2 DL, but 3 of these can be only be represented individually, but not collectively~\footnote{These three axioms refer to the irreflexivity, asymmetry and transitivity of \textit{proper part of}, because OWL's global restrictions  prohibit the combination of these characteristics for the same object property. 
}.

\begin{figure}
\centering
\begin{minipage}{.5\textwidth}
\flushleft
  \includegraphics[width=.9\linewidth]{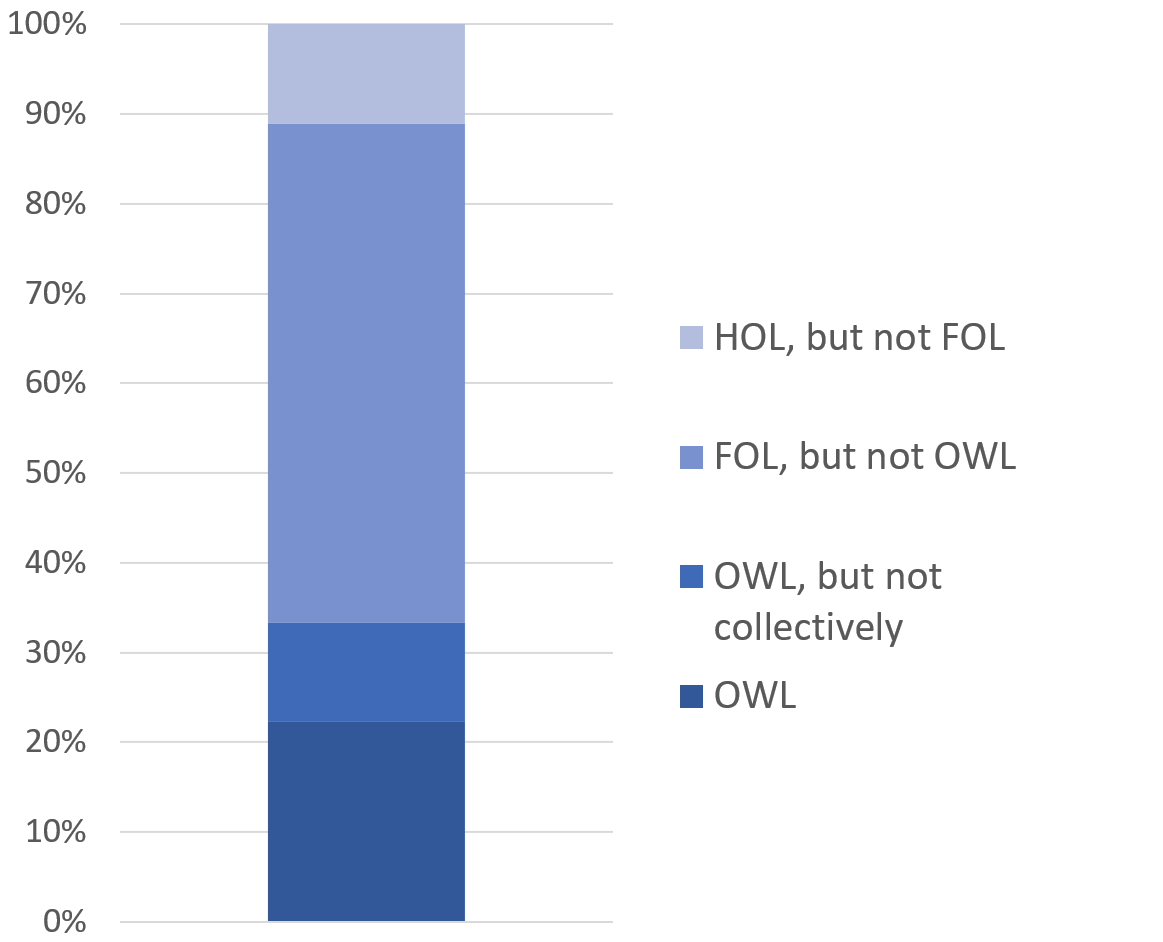}
	\caption{The shares of KGEMT axioms which can be\\ represented in OWL, FOL and HOL respectively.}
  \label{fig:evalkgemtdiagram}
\end{minipage}%
\begin{minipage}{.5\textwidth}
  \flushright
  {\setstretch{.9}
\begin{lstlisting}[label=lst:kgemt4p, caption=The premise ontology of a test case based on a KGEMT axiom., linerange={8-20}]
Class: ovgu_part
    EquivalentTo:
        part_of value ovgu
    SubclassOf:
        overlap value magdeburg

ObjectProperty: overlap
ObjectProperty: part_of
    Annotations:
        tptp_annotation "![X, Y]: (~part_of(Y, X) => ?[Z]: (part_of(Z, Y) & ~overlap(Z, X)))"

Individual: ovgu
Individual: magdeburg
\end{lstlisting}
}
\end{minipage}
\end{figure}

For the purposes of our evaluation, we can use the set of 27 axioms  in \cite{keet2012ontoparts}  as a basis for generating test cases.
Out of the 27 axioms, only 3 are not representable in FOL, but require Higher Order Logic (HOL).
Thus, nearly 90\% of the axioms are representable in either OWL or FOL (see Figure~\ref{fig:evalkgemtdiagram}). 
The 15 axioms which are representable in FOL, but not in OWL, were used for creating test cases. An additional test case is based on the axioms which cannot be represented collectively in OWL, leading to a total of 16 test cases.

Each test case consists of a FOWL premise ontology which consists mostly of OWL axioms but includes one of the KGEMT axioms as a FOL annotation, and an OWL conjecture ontology, which cannot be inferred by OWL reasoning from the premise ontology. 
For example, the premise ontology in Listing~\ref{lst:kgemt4p} 
contains OWL axioms that express that every part of OVGU is part of Magdeburg. Further, it includes a KGEMT axiom as a TPTP annotation, which states:  for all x and y, if y (e.g., OVGU) is not part of x (e.g., Magdeburg), there has to be an instance z that is part of y, but does not overlap x.
The conjecture ontology of this example states that OVGU is part of Magdeburg. 

%
%
In each of the 16 test cases {\gavelowl} is able to prove that the conjecture ontology is entailed from the premise ontology. 
%
%
This result illustrates that though some axioms had to be left out because they are not representable in FOL, the axiomatisation of an ontology covering mereotopological concepts can become significantly stronger when using FOL annotations. More generally, the results of both test cases indicate  our tool is working correctly and that it is able to integrate the FOL annotations and the ontology's OWL part into a coherent logical theory that can be used for reasoning tasks.
Further, they illustrate that the additional expressivity granted by FOL 
annotations may be used to address some of the limitations of OWL that have been discussed in the literature. 
%
%
%
All test cases can be found on the project's GitHub page~\footnote{\url{https://github.com/gavel-tool/python-gavel-owl}}.

\subsection{Extending BFO}
With FOWL, we can not only add individual FOL axioms to OWL ontologies, but also integrate whole FOL and OWL ontologies. 
This is especially useful where OWL domain ontologies are based on FOL upper level ontologies.
To demonstrate this, we used the BFO in combination with the Ontology for Biomedical Investigations (OBI)~\cite{bandrowski2016obi}.
OBI uses BFO 2.0 OWL, which due to restrictions in the OWL language does not represent the complete axiomatisation of BFO. 
The latter is only available in FOL.
Thus, it is not able to check if OBI fully conforms to BFO with an OWL reasoner.

BFO 2.0 OWL does however already contain annotations containing axioms from BFO 2.0 FOL in CLIF syntax.
We used these annotations to build a new BFO version where the annotations are interpretable by CLIF parsers and can be combined with BFO 2.0 OWL as well as OWL domain ontologies.
For that, we had to make several changes: 
Firstly, the annotations in BFO 2.0 OWL contain additional comments besides the CLIF axioms which we removed. 
Secondly, the axioms in BFO 2.0 FOL are temporalized, leading to ternary predicates which are not compatible with OWL ontologies. 
Therefore, we removed this temporalization. 
This is based on the assumption that OWL ontologies in general make time-independent statements.

Additionally to the FOL axioms, our version of BFO contains OWL object properties from the Relation Ontology (RO)~\footnote{https://github.com/oborel/obo-relations/} and BFO 2020 OWL~\footnote{https://github.com/BFO-ontology/BFO-2020}, another BFO version, corresponding to the binary predicates in our modified BFO 2.0 FOL axioms. 
This is necessary because BFO 2.0 OWL does not contain any object properties, just the BFO classes.
Where possible, we used the object properties from RO, because they are already used by many BFO-based ontologies, including OBI. In addition, we have adapted the names of binary FOL predicates to match the object properties from RO and BFO 2020 OWL where necessary.
This ensures that {\gavelowl} can match both types of relations, FOL predicates and OWL object properties, correctly.
Similarly, for the unary predicates, we replaced the BFO CLIF labels with the BFO OWL labels as they are specified in BFO 2.0 OWL.
Lastly, we imported our BFO version\footnote{The FOWL version of BFO can be accessed at \url{https://github.com/gavel-tool/BFOnow}} into OBI and used {\gavelowl} to translate the resulting ontology to FOL. 
Subsequently, we checked the FOL theory for satisfiability, which correlates with the ontology's internal consistency. 
Internal consistency means that a model can be constructed for the ontology, compared to external consistency, which signifies that a model can be constructed from datasets that represent an ontology's intended use~\cite{stephen2020modelfinding}.
We checked the latter one by adding one individual to every OBI-class and repeating the process. 
Here, the intended use case we considered is one where all classes can be instantiated.

In total, we were able to detect 4 inconsistencies, 2 of them external inconsistencies.
For illustration purposes we discuss one of these inconsistencies in more detail.
It arises from a conflict mainly between two axioms, one being BFO 2.0 FOL axiom number 134-001, 
\begin{verbatim}
(forall (x) (if ("independent continuant" x) 
                (exists (r) (and ("spatial region" r) ("located in" x r))))),
\end{verbatim}
stating that for every \textit{independent continuant}, there is a \textit{spatial region} in which it is located.
It is in conflict with an OWL axiom belonging to the OBI class \textit{local field potential recording},
\begin{verbatim}
(has_specified_input some 
    ((tissue and ('located in' only brain))
        and ('has role' some 'evaluant role'))),
\end{verbatim}
which ascertains that a \textit{local field potential recording} is related to an instance of \textit{tissue} that only has a \textit{located in}-relationship to instances of \textit{brain}.
Since \textit{tissue} is an \textit{independent continuant}, every instance of it has to be located in some \textit{spatial region}, which is an \textit{immaterial entity}.
\textit{immaterial entity} is disjoint with \textit{material entity} of which \textit{brain} is a subclass.
Thus, no instance of \textit{tissue} can be \textit{located in only brain} and \textit{local field potential recording} cannot be instantiated.

This example illustrates that {\gavel} can be used to find inconsistencies in existing OWL ontologies that have previously been undetected because of the lack of access to automatic reasoning with FOL axioms.

While using the BFO FOL axioms, we also noticed an error in axiom [062-002], where the arguments of the bearer-of relation have been swapped. 
This indicates that using automatic reasoning can help detect mistakes both in FOL and OWL ontologies which would have remained unnoticed if the languages had not been connected.

Our BFO version can be applied not only to OBI, but to other BFO-based ontologies as well. 
It and a version containing only the modified BFO 2.0 FOL axioms and the object properties can be found on GitHub~\footnote{\url{https://github.com/gavel-tool/BFOnow}}.

\subsection{Application to the Chemistry Ontology ChEBI} 

\begin{figure}[tbhp]
  \parbox{.6\textwidth}{
    \centering
    \includegraphics[width=0.6\textwidth]{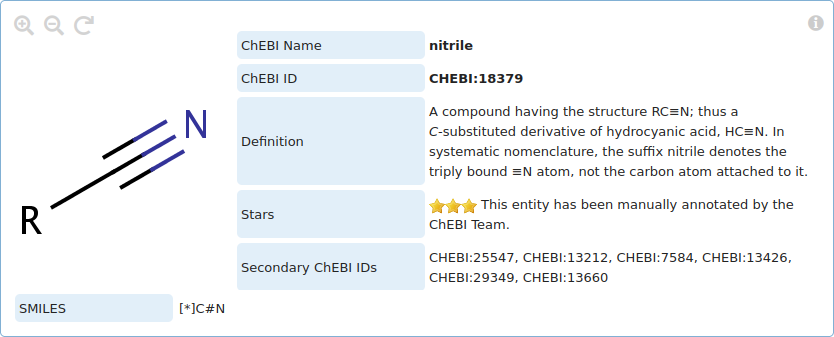}
    \caption{The class \textit{nitrile} as represented on the ChEBI website.}
    \label{fig:nitrile}
  } \hfill
  \parbox{.3\textwidth}{
    \centering
    \includegraphics[width=.3\textwidth]{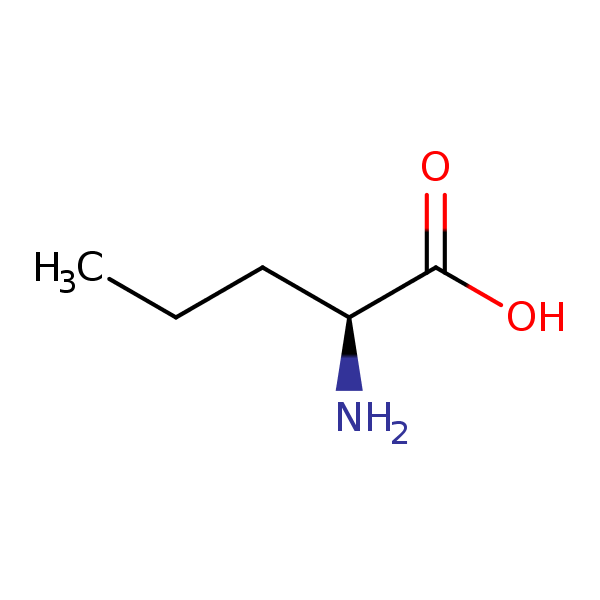}
    \caption{The class \textit{L-2-aminopentanoic acid}, which had been wrongly axiomatised as a subclass of nitrile.}
    \label{fig:l-2-aminopentanoic-acid}
  }
\end{figure}

ChEBI \cite{hastings2016chebi} is a freely accessible, open OWL ontology for a biologically relevant chemical entities. This mainly includes substances that occur naturally in biological processes or synthetic substances that interact with these processes. 
 Figure~\ref{fig:nitrile} shows an example of a ChEBI class with annotations. In addition to the molecular graph and textual descriptions and definitions, ChEBI contains a representation of the molecular class as SMILES specification. 
 SMILES is a  representation language that is specially developed for molecular entities, and it allows the direct serialisation of a molecular structure.
ChEBI is the largest publicly available ontology for chemical entities and covers, as of August 2022, a total of more than 180,000 classes.
The maintenance and further development of such an ontology requires a large, heterogeneous team of ontology and domain experts. To facilitate this process, ChEBI allows third-party annotation and uses a special rule-based automatic classifier, ClassyFire \cite{djoumbou_feunang_classyfire_2016}. However, ChEBI is generally quite weakly axiomatised and, to our knowledge, no symbolic methods beyond the standard OWL-DL reasoning are currently used to validate ChEBI.

For our  case study we extended ChEBI with more than 129,193 FOL annotations\footnote{
The FOWL extension of ChEBI that contains the these additional annotations may be accessed at \url{https://doi.org/10.5281/zenodo.7038560} }. 
 and used these to  validate ChEBI's OWL axioms. 
Most of these FOL axioms were automatically generated based on SMILES annotations. For the purpose of this case study, we consider two different categories of classes of chemical entities in ChEBI. 

The first category contains classes that are associated with SMILES strings that contain unspecified parts. 
 These unspecified parts are represented in the SMILES strings as \textit{'*'} and in the molecular graph as \textit{R}, representing the attachment at a particular point of a group or sub-structural component. 
 For example, the class \textit{Nitrile} in Figure~\ref{fig:nitrile} is incompletely specified, because the ``*'' might be replaced by a lot of different structures. 
 
In contrast, there are the classes of chemical entities whose structure  is completely specified by their SMILES string. 
These classes occur primarily at the lowest levels of the ontological hierarchy and usually have no subclasses, and if they do, they represent stereoisomers, i.e. the subclasses only differ in spacial orientation and not in structure. 
 For example, 
  \textit{L-2-aminopentanoic acid} in Figure~\ref{fig:l-2-aminopentanoic-acid} is fully specified, i.e. all molecules that instantiate this class have the same molecular structure. 

The difference between these two categories of classes is important, since they lead to different kinds of FOL axioms. 

Classes that are associated with an incomplete SMILES specification are typically defined via the presence of some substructure. For example, Listing~\ref{lst:nitrile} shows an axiomatisation for the molecular class \textit{nitrile},  automatically generated based on its SMILES annotation. (Note that the carbon atom that is bound to the nitrogen is explicitly represented.)
\begin{lstfloat}
   \begin{lstlisting}[caption=First order axiomatisation for the molecular class nitrile,numbers=none,label=lst:nitrile]
  fof(chebi_18379_0, 
    axiom, 
    ![M]:(chebi18379(M) <=> (
      ?[N2, N1]:(  c(N1) & part_of(N1, M) & has_no_charge(N1) 
                 & n(N2) & part_of(N2, M) & has_no_charge(N2)
                 & has_triple_bond_to(N1, N2) & N2 != N1
                 & connected(M))))).
  \end{lstlisting}
\end{lstfloat}

In addition, the SMILES annotations of this category of classes often only provide an incomplete representation of the textual definitions available in ChEBI. 
For example, consider medium- and long-chain fatty acids (CHEBI:59554, CHEBI:15904). Their smiles annotations are identical, but their definitions distinguish them by their respective chains lengths. Medium-chain fatty acids have chains that consist of at least 6 carbon atoms, but not more than 12. Long chains consist of at least 13, but less than 22 carbon atoms. Neither smiles nor OWL DL can express such structures. Another example is classes that are not just defined by the presence but also by the absence of some substructures, since SMILES has no equivalent to a negation. In some cases where we noticed that there were mismatches or omissions in the generated FOL class definitions, we  adjusted the FOL axioms manually. 


The formalisation of the second category of classes, i.e., the fully specified ones, is different in two ways. Since all instances of such a class have the same molecular structure, we represent them as a `prototypical instance'. Hence, their formalisations do not involve quantifiers but individuals. 
From a  logical point of view this is not necessary (we could follow the same pattern of definition as with the other type of classes), but it  improves the performance of automatic reasoning significantly. 
In addition, since the SMILES specification is complete, we need to express that the given molecule contains no other parts than the ones explicitly mentioned in the axioms.   
Hence, the formalisation of these classes contain a formula of the form  
of the form \verb/![X]:partOf(X, mol) => (X=n1 | X=n2 | ...)/.  For example, the FOL axiom for water (CHEBI:15377) is represented in Listing~\ref{lst:water}.

\begin{lstfloat}
\begin{lstlisting}[caption=First order axiomatisation for the `prototypical instance' of water,numbers=none,label=lst:water]
fof(chebi_15377_inst, 
    axiom, 
        o(n15377_0) & part_of(n15377_0, m15377) & has_no_charge(n15377_0)
      & h(n15377_1) & part_of(n15377_1, m15377) & has_no_charge(n15377_1) 
      & h(n15377_2) & part_of(n15377_2, m15377) & has_no_charge(n15377_2) 
      & has_single_bond_to(n15377_0, n15377_1)
      & has_single_bond_to(n15377_0, n15377_2)
      & ~has_bond(n15377_1, n15377_2)
      & ![X]:(part_of(X, m15377)
               => (X=n15377_0 | X=n15377_1 | X=n15377_2))
      & connected(m15377)). 
\end{lstlisting}
\end{lstfloat}

For our experiments we used our extension of ChEBI with these FOL annotations together with an additional ontology that added some chemical background knowledge (e.g., that chemical elements are disjoint). Using {\gavel} we proved that the current version of ChEBI was, in fact, inconsistent. We therefore decided to investigate further and split the ontology into smaller fragments.

For a detailed analysis, we selected a subset of ChEBI consisting of 80 incompletely specified classes and 6569 completely specified classes, 
 each of  which is represented by some `prototypical individual'.  For each of these `prototypical individuals' $a$ and each incompletely specified class $B$, we created the conjecture $B(a)$. Because $a$ is structurally indistinguishable from all other members of its class, the conjecture is $B(a)$ equivalent to the assertion that the class of $a$ is a subclass of $B$.  

 We used {\gavel} and Vampire to attempt to prove these conjectures and compared the results to the transitive closure of subsumption relations in ChEBI. Assuming that all information in ChEBI is correct, a conjecture $B(a)$ should only be provable from the FOL theory,  if $a \sqsubseteq B$ is true in ChEBI. Furthermore, Vampire should only be able to find a counterexample, if $a \sqsubseteq B$ is not an axiom in ChEBI.

The experiments were conducted on a CPU clusters with 80 jobs, using 4 CPUs each. Each job focussed on a single molecular class and attempted to prove membership of all 6569 instances individually, limited by a time out of 30 seconds per proof. Each of these jobs terminated within 24 hours, while 31.06\% of all proof attempts were stopped by the time-out.
Our experiments showed that the proof complexity scales directly with the complexity of the molecule class. Some complex classes, such as \textit{isoflavones} (CHEBI:38757), exceeded the time-out of 30s on all instances. While the results of the FOL reasoning were mostly consistent with the OWL axioms, there were some interesting findings, which can be distinguished in two categories: 
\begin{description}
\item[Unexpected proof] $a \sqsubseteq B$ is not an axiom in ChEBI. Yet, $B(a)$ was provable based on the FOL annotations.  This occurred in 10132 proof attempts (0.0279\% of all proof attempts).
\item[Unexpected counter-example] $a \sqsubseteq B$ is an axiom in ChEBI. Yet, Vampire refuted the conjecture $B(a)$. This occurred in 256 proof attempts (0.0007\% of all proof attempts).
\end{description}

Unexpected proofs are likely to originate from  SMILES annotations that do not specify a class adequately.  In these cases the FOL definitions that are automatically generated are too wide, which leads to unexpected proofs.    
Thus, each of these unexpected proofs is an indication that either a SMILES annotation in ChEBI is erroneous or that SMILES is not expressive enough to represent the definition of the class (e.g., because the class is partially defined by the absence of a feature). 
As discussed above, in some cases we addressed these unexpected proofs by changing the FOL definitions manually.

Unexpected counter-examples are candidates for inconsistencies in the underlying ontology. For example, the molecule in Figure~\ref{fig:l-2-aminopentanoic-acid} was classified as a nitrile even though it has a single tetrahedral bond between the carbon atom and the sodium atom and not the triple bond required for nitriles. We reported this finding to the ChEBI developers and it has been acknowledged as an error and fixed in the ontology\footnote{\url{https://github.com/ebi-chebi/ChEBI/issues/4281}}.

Thus, this case study illustrates that FOL annotations may be used to  represent knowledge in a large domain ontology, and that reasoning with these FOL annotations with {\gavel} can be successfully used to identify errors in such an ontology.


\section{Related Work}

There are several works on extending the expressivity of OWL ontologies that are closely related to our work. In \cite{schneider2012owlnorestrictions} Schneider et al.  present 12 useful modelling patterns that are not expressible  in OWL 2 DL, because they violate its global restrictions (e.g., 
axiomatising an object property as strict partial order).  
For evaluation purposes these modelling patterns are implemented in OWL ontologies and also translated into FOL. 
Their experiments show that, while all of the OWL reasoners failed on these test cases, the FOL reasoners were able to draw the desired inferences.  
Hence, the authors illustrated the feasibility to  FOL reasoning for OWL ontologies. 
However,  \cite{schneider2012owlnorestrictions}  only  considers ontologies that consist of OWL constructs, while our work enables the use of arbitrary FOL annotations as part of OWL ontologies. In that sense our work is more general.  

Another important strategy to extend the expressivity of OWL is by adding rules. 
This is supported by the W3C recommendation  Rule Interchange Format (RIF)\cite{kif2013}, which supports a variety of dialects based on different logics (e.g., Horn Logic or Datalog), but none of them supports full FOL. 
Significantly older, but influential is SWRL~\cite{horrocks2004swrl}, which combines the OWL and OWL Lite sublanguages with features of the RuleML~\cite{boley2001ruleML}.
It uses Horn-like rules in the form of implications between an antecedent called the body and a consequent called the head. However, the extension of OWL by SWRL is not decidable. 
For this reason Motik et. al.~\cite{motik2005query} suggest an alternative, 
which combines OWL DL with DL-safe rules. These are, roughly speaking, function-free Horn rules, where variables are bound only to named individuals.   
%
Even though the combination of OWL and SWRL has a higher expressivity than OWL, it does not provide the same expressivity as FOL.
For this reason it was proposed to extend
SWRL by arbitrary FOL axioms \cite{patel2005swrlfol}. However, as far as we are aware, this proposal has never been implemented.

%
%

The Distributed Ontology, Modeling, and Specification Language (DOL)~\cite{mossakowski2015dol,omg2018dol} is an Object Management Group (OMG) specification, which -- among many features -- supports the representation of heterogeneous ontologies and translation between ontology languages. 
DOL is implemented by Hets~\cite{mossakowski2007hets, mossakowski2018hets} and supports, in particular, OWL, CLIF, and TPTP. Thus, many capabilities of {\gavel} are also provided by Hets. However, a major difference is the fact that DOL/Hets assumes that a heterogeneous ontology is composed of different sub-ontologies which are written in only one language, e.g., an DOL ontology might import an  OWL file and a separate CLIF file. This solution is inconvenient for ontology developers, since the heterogeneous ontology might not be edited by established editors like Protégé and axioms concerning one entity might be spread over different files. Further, {\gavel} provides the capability to automatically align the signatures of the various ontologies. Another difference is that {\gavel} is implemented in Python and is more lightweight than Hets.

\section{Importance, Impact and Conclusions}

As discussed in section~\ref{sec:introduction}, the 
field of Applied Ontology is divided by a language barrier. 
Since ontologists who work on foundational issues typically prefer more expressive languages than OWL, the result of their research cannot be directly applied to ontologies that are written in OWL -- which includes the vast majority of all domain ontologies. In particular this is illustrated by the fact that top-level ontologies such as BFO and DOLCE are represented primarily in FOL, but the versions that are actually used by domain ontologies are in OWL. 
Thus, so far it was not possible to check whether these domain ontologies actually conform to their top-level ontology. 

{\gavel} is designed to help ontology developers to bridge the divide, since it enables the development of FOWL ontologies, i.e., OWL ontologies with FOL annotations. 
In section~\ref{sec:evaluation} we have illustrated our approach by validating the 
Ontology for Biomedical Investigations (OBI).
OBI is a relatively small  (about 5000 entities), but widely used  ontology\footnote{Between July 2021 and July 2022 OBI was downloaded 2163 times from Bioportal. \url{https://bioportal.bioontology.org/ontologies/OBI}}, which uses the (OWL version of) BFO as its top-level ontology.  
By combining OBI with FOL annotations that were derived from FOL annotations published as part of BFO 2.0 OWL, we were able to detect several inconsistencies.
This method of  evaluation is not limited to OBI, but is applicable to any of the more than 250 domain ontologies that use BFO as their top-level ontology. By creating analogous FOWL versions of other top-level ontologies, the same strategy may be applied to them as well. 

Furthermore, we discussed how {\gavel} is able to reason with a FOWL ontology that extends an OWL ontology with complex mereological axioms. Mereology is one of the best studied areas in applied ontology, but OWL is not expressive enough to represent many mereological distinctions. This example illustrates that {\gavel} enables ontology developers to enhance OWL ontologies with the results of research on foundational ontologies.  

The benefit of reasoning with FOWL ontologies with Gavel is not limited to the integration of foundational and domain ontologies. As our third use case illustrates, FOL annotations may be fruitfully used to express domain knowledge, which cannot be expressed in OWL -- even in the case  of very large ontologies like ChEBI. By applying automatic reasoning we were able to detect a number of potential problems in ChEBI, of which we reported two and they have already been fixed by ChEBI developers. 
A surprising result is the fact that FOL reasoning with ChEBI is a viable option, even though ChEBI is a very large ontology. Obviously, there were time-outs, but a large majority of proof attempts returned a result and, thus, illustrated the usefulness of the approach.   

In conclusion, in this paper we have shown the benefits of FOWL ontologies and presented the tool {\gavel}, which is able to support reasoning with these heterogeneous ontologies. 
Since these ontologies are syntactically valid OWL ontologies, using these annotations does not break existing tool chains, and, thus, the barrier for using this approach is low. 

In the future we are planning to extend our approach from FOL to higher-order logic. This would enable us to cover the mereological axioms in \cite{keet2012ontoparts} that we had to exclude from our evaluation in section~\ref{sec:evaluation}. 
Further, there are definitions of chemical classes that require monadic second-order logic  \cite{kutz2012modelling}. 

Another area we are planning to explore is the development of a more densely axiomatised BFO FOWL ontology. 
For the purposes of this paper we have mainly considered axioms that have been derived from the FOL annotations that are already present in BFO 2.0 OWL.  However, these axioms are relatively sparse and could be significantly improved. In combination with Gavel this would enable a  better validation of domain ontologies that are based on BFO.   

Further, we are planning to integrate our workflow for a logical evaluation of chemical subclass relationships with our work on automatic ontology
 extension \cite{interpretable_ontology_extension_2022,hastings2021learning,memariani2021}. 
In these works we used machine learning techniques to predict for a given unknown chemical molecule its suitable classification in ChEBI. From a logical point of view, we trained models that -- when provided with the structure of a chemical molecule -- are able to generate OWL subsumption conjectures. We have already used OWL disjointness axioms to check for some errors in these conjectures. 
In the future we are planning to use FOL annotations of ChEBI to further automatically validate these conjectures with logical reasoning.     
 






\bibliographystyle{plain}           
\bibliography{fowl_swj}        

\begin{thebibliography}{10}

\bibitem{aiello2007spatiallogics}
Marco Aiello, Ian Pratt-Hartmann, Johan Van~Benthem, et~al.
\newblock {\em {Handbook of Spatial Logics}}, volume~4.
\newblock Springer, 2007.

\bibitem{bandrowski2016obi}
Anita Bandrowski, Ryan Brinkman, Mathias Brochhausen, Matthew~H Brush, Bill
  Bug, Marcus~C Chibucos, Kevin Clancy, M{\'e}lanie Courtot, Dirk Derom, Michel
  Dumontier, et~al.
\newblock {The Ontology for Biomedical Investigations}.
\newblock {\em PloS one}, 11(4):e0154556, 2016.

\bibitem{boley2001ruleML}
Harold Boley, Said Tabet, and Gerd Wagner.
\newblock {Design Rationale for RuleML: A Markup Language for Semantic Web
  Rules}.
\newblock In {\em SWWS}, volume~1, pages 381--401, 2001.

\bibitem{borgo2022Foundational}
Stefano Borgo, Antony Galton, and Oliver Kutz.
\newblock Special issue: Foundational ontologies in action.
\newblock {\em Applied Ontology}, 17-1:1--16, 2022.

\bibitem{djoumbou_feunang_classyfire_2016}
Yannick Djoumbou~Feunang, Roman Eisner, Craig Knox, Leonid Chepelev, Janna
  Hastings, Gareth Owen, Eoin Fahy, Christoph Steinbeck, Shankar Subramanian,
  Evan Bolton, Russell Greiner, and David~S. Wishart.
\newblock {ClassyFire}: automated chemical classification with a comprehensive,
  computable taxonomy.
\newblock {\em Journal of Cheminformatics}, 8(1):61, December 2016.

\bibitem{flugel2021fowl}
Simon Fl{\"u}gel, Anna Kleinau, Fabian Neuhaus, Martin Glauer, and Janna
  Hastings.
\newblock {FOWL--An OWL to FOL Translator}.
\newblock {\em Proceedings of the Joint Ontology Workshops 2021}, 2021.

\bibitem{iso2018commonlogic}
International~Organization for Standardization.
\newblock {Common Logic (CL) - A Framework for a Family of Logic-based
  Languages}.
\newblock Standard, International Organization for Standardization, Geneva, CH,
  07 2018.

\bibitem{interpretable_ontology_extension_2022}
Martin Glauer, Adel Memariani, Fabian Neuhaus, Till Mossakowski, and Janna
  Hastings.
\newblock Interpretable ontology extension in chemistry.
\newblock {\em Semantic Web Journal}, accepted.

\bibitem{glimm2014hermit}
Birte Glimm, Ian Horrocks, Boris Motik, Giorgos Stoilos, and Zhe Wang.
\newblock {HermiT}: an {OWL} 2 {R}easoner.
\newblock {\em Journal of Automated Reasoning}, 53(3):245--269, 2014.

\bibitem{omg2018dol}
Object~Management Group.
\newblock {Distributed Ontology, Model, and Specification Language}.
\newblock Standard, Object Management Group, Milford, MA, USA, 03 2018.

\bibitem{hastings2021learning}
Janna Hastings, Martin Glauer, Adel Memariani, Fabian Neuhaus, and Till
  Mossakowski.
\newblock Learning chemistry: exploring the suitability of machine learning for
  the task of structure-based chemical ontology classification.
\newblock {\em Journal of Cheminformatics}, 13(1):1--20, 2021.

\bibitem{hastings2016chebi}
Janna Hastings, Gareth Owen, Adriano Dekker, Marcus Ennis, Namrata Kale,
  Venkatesh Muthukrishnan, Steve Turner, Neil Swainston, Pedro Mendes, and
  Christoph Steinbeck.
\newblock Chebi in 2016: Improved services and an expanding collection of
  metabolites.
\newblock {\em Nucleic acids research}, 44(D1):D1214--D1219, 2016.

\bibitem{horridge2011owlapi}
Matthew Horridge and Sean Bechhofer.
\newblock {The OWL API: A Java API for OWL Ontologies}.
\newblock {\em Semantic Web}, 2(1):11--21, 2011.

\bibitem{owl2directsemantics}
Ian Horrocks, Bijan Parsia, and Uli Sattler.
\newblock {OWL 2 Web Ontology Language Direct Semantics (Second Edition)},
  2012.

\bibitem{horrocks2004swrl}
Ian Horrocks, Peter~F Patel-Schneider, Harold Boley, Said Tabet, Benjamin
  Grosof, Mike Dean, et~al.
\newblock {SWRL: A semantic web rule language combining OWL and RuleML}.
\newblock {\em W3C Member Submission}, 21(79):1--31, 2004.

\bibitem{keet2012ontoparts}
C~Maria Keet, Francis~C Fern{\'a}ndez-Reyes, and Annette Morales-Gonz{\'a}lez.
\newblock {Representing Mereotopological Relations in OWL Ontologies with
  OntoPartS}.
\newblock In {\em Extended Semantic Web Conference}, pages 240--254. Springer,
  2012.

\bibitem{kif2013}
Michael Kifer and Harold Boley.
\newblock Rif overview (second edition).
\newblock Technical report, World Wide Web Consortium (W3C), 2013.

\bibitem{kovacs2013vampire}
Laura Kov{\'a}cs and Andrei Voronkov.
\newblock {First-order Theorem Proving and Vampire}.
\newblock In {\em International Conference on Computer Aided Verification},
  pages 1--35. Springer, 2013.

\bibitem{kutz2012modelling}
Oliver Kutz, Janna Hastings, and Till Mossakowski.
\newblock Modelling highly symmetrical molecules: Linking ontologies and
  graphs.
\newblock In {\em International Conference on Artificial Intelligence:
  Methodology, Systems, and Applications}, pages 103--111. Springer, 2012.

\bibitem{mcmurry_identifiers_2017}
Julie~A McMurry, Nick Juty, Niklas Blomberg, Tony Burdett, Tom Conlin, Nathalie
  Conte, Mélanie Courtot, John Deck, Michel Dumontier, Donal~K Fellows,
  Alejandra Gonzalez-Beltran, Philipp Gormanns, Jeffrey Grethe, Janna Hastings,
  Jean-Karim Hériché, Henning Hermjakob, Jon~C Ison, Rafael~C Jimenez, Simon
  Jupp, John Kunze, Camille Laibe, Nicolas Le~Novère, James Malone,
  Maria~Jesus Martin, Johanna~R McEntyre, Chris Morris, Juha Muilu, Wolfgang
  Müller, Philippe Rocca-Serra, Susanna-Assunta Sansone, Murat Sariyar,
  Jacky~L Snoep, Stian Soiland-Reyes, Natalie~J Stanford, Neil Swainston,
  Nicole Washington, Alan~R Williams, Sarala~M Wimalaratne, Lilly~M Winfree,
  Katherine Wolstencroft, Carole Goble, Christopher~J Mungall, Melissa~A
  Haendel, and Helen Parkinson.
\newblock Identifiers for the 21st century: {How} to design, provision, and
  reuse persistent identifiers to maximize utility and impact of life science
  data.
\newblock {\em PLoS Biology}, 15(6):e2001414, June 2017.

\bibitem{memariani2021}
Adel Memariani, Martin Glauer, Fabian Neuhaus, Till Mossakowski, and Janna
  Hastings.
\newblock Automated and explainable ontology extension based on deep learning:
  A case study in the chemical domain.
\newblock In Roberto Confalonieri, Oliver Kutz, and Diego Calvanese, editors,
  {\em International Workshop on Data meets Applied Ontologies in Explainable
  AI (DAO-XAI 2021)}, volume 2998 of {\em CEUR Workshop Proceedings},
  http://ceur-ws.org/Vol-2998/, 2021.

\bibitem{mossakowski2018hets}
Till Mossakowski and Mihai Codescu.
\newblock {The Heterogeneous Tool Set --- some Recent Developments and
  Highlights}.
\newblock {\em 24th International Workshop on Algebraic Development Techniques
  2018}, 2018.

\bibitem{mossakowski2015dol}
Till Mossakowski, Mihai Codescu, Fabian Neuhaus, and Oliver Kutz.
\newblock {The Distributed Ontology, Modeling and Specification Language--DOL}.
\newblock In {\em The Road to Universal Logic}, pages 489--520. Springer, 2015.

\bibitem{mossakowski2007hets}
Till Mossakowski, Christian Maeder, and Klaus L{\"u}ttich.
\newblock The {H}eterogeneous {T}ool {S}et, {Hets}.
\newblock In {\em International Conference on Tools and Algorithms for the
  Construction and Analysis of Systems}, pages 519--522. Springer, 2007.

\bibitem{motik2005query}
Boris Motik, Ulrike Sattler, and Rudi Studer.
\newblock {Query Answering for OWL-DL with Rules}.
\newblock {\em Journal of Web Semantics}, 3(1):41--60, 2005.

\bibitem{patel2005swrlfol}
Peter~F Patel-Schneider.
\newblock {A Proposal for a SWRL Extension Towards First-order Logic}.
\newblock {\em W3C Member Submission, April}, 2005.

\bibitem{schneider2012owlnorestrictions}
Michael Schneider, Sebastian Rudolph, and Geoff Sutcliffe.
\newblock Modeling in {OWL} 2 without {R}estrictions.
\newblock In Mariano Rodriguez{-}Muro, Simon Jupp, and Kavitha Srinivas,
  editors, {\em Proceedings of the 10th International Workshop on {OWL:}
  Experiences and Directions {(OWLED} 2013) co-located with 10th Extended
  Semantic Web Conference {(ESWC} 2013), Montpellier, France, May 26-27, 2013},
  volume 1080 of {\em {CEUR} Workshop Proceedings}. CEUR-WS.org, 2013.

\bibitem{stephen2020modelfinding}
Shirly Stephen and Torsten Hahmann.
\newblock {Model-Finding for Externally Verifying FOL Ontologies: A Study of
  Spatial Ontologies}.
\newblock {\em Frontiers in Artificial Intelligence and Applications},
  330:233--248, 2020.

\bibitem{sutcliffe2006tptp}
Geoff Sutcliffe, Stephan Schulz, Koen Claessen, and Allen Van~Gelder.
\newblock {Using the TPTP Language for Writing Derivations and Finite
  Interpretations}.
\newblock In {\em International Joint Conference on Automated Reasoning}, pages
  67--81. Springer, 2006.

\bibitem{sutcliffe2017tptp}
Geoffrey Sutcliffe.
\newblock {The TPTP Problem Library and Associated Infrastructure: From CNF to
  TH0, TPTP v6. 4.0}.
\newblock {\em Journal of Automated Reasoning}, pages 1--20, 2017.

\end{thebibliography}

%

\end{document}